\def\year{2019}\relax
\documentclass[letterpaper]{article} 
\usepackage{aaai19}  
\usepackage{times}  
\usepackage{helvet}  
\usepackage{courier}  
\usepackage{url}  
\usepackage{graphicx}  
\usepackage{booktabs}
\usepackage{array}
\usepackage{amsmath}
\usepackage{tikz}
\usepackage{amssymb}
\usepackage{color,colortbl}
\usepackage{multirow}
\usepackage{url}
\usepackage[T1]{fontenc}
\newcommand{\utterance}[1]{``#1''}

\title{A Natural Language Corpus of Common Grounding under \\ Continuous and Partially-Observable Context}
\author{Takuma Udagawa\\
{The University of Tokyo, Tokyo, Japan}\\
{takuma\_udagawa@nii.ac.jp}
\And
Akiko Aizawa\\
{National Institute of Informatics, Tokyo, Japan}\\
{aizawa@nii.ac.jp}
}

\begin{document}
\maketitle
\begin{abstract}
Common grounding is the process of creating, repairing and updating mutual understandings, which is a critical aspect of sophisticated human communication. However, traditional dialogue systems have limited capability of establishing common ground, and we also lack task formulations which introduce natural difficulty in terms of common grounding while enabling easy evaluation and analysis of complex models. In this paper, we propose a minimal dialogue task which requires advanced skills of common grounding under continuous and partially-observable context. Based on this task formulation, we collected a largescale dataset of 6,760 dialogues which fulfills essential requirements of natural language corpora. Our analysis of the dataset revealed important phenomena related to common grounding that need to be considered. Finally, we evaluate and analyze baseline neural models on a simple subtask that requires recognition of the created common ground. We show that simple baseline models perform decently but leave room for further improvement. Overall, we show that our proposed task will be a fundamental testbed where we can train, evaluate, and analyze dialogue system's ability for sophisticated common grounding.
\end{abstract}

\section{Introduction}
\label{section:introduction}

\begin{figure}[ht]
\begin{tikzpicture}
\node[inner sep=0pt] (agent_0) at (0,0)
  {\includegraphics[width=0.48\columnwidth]{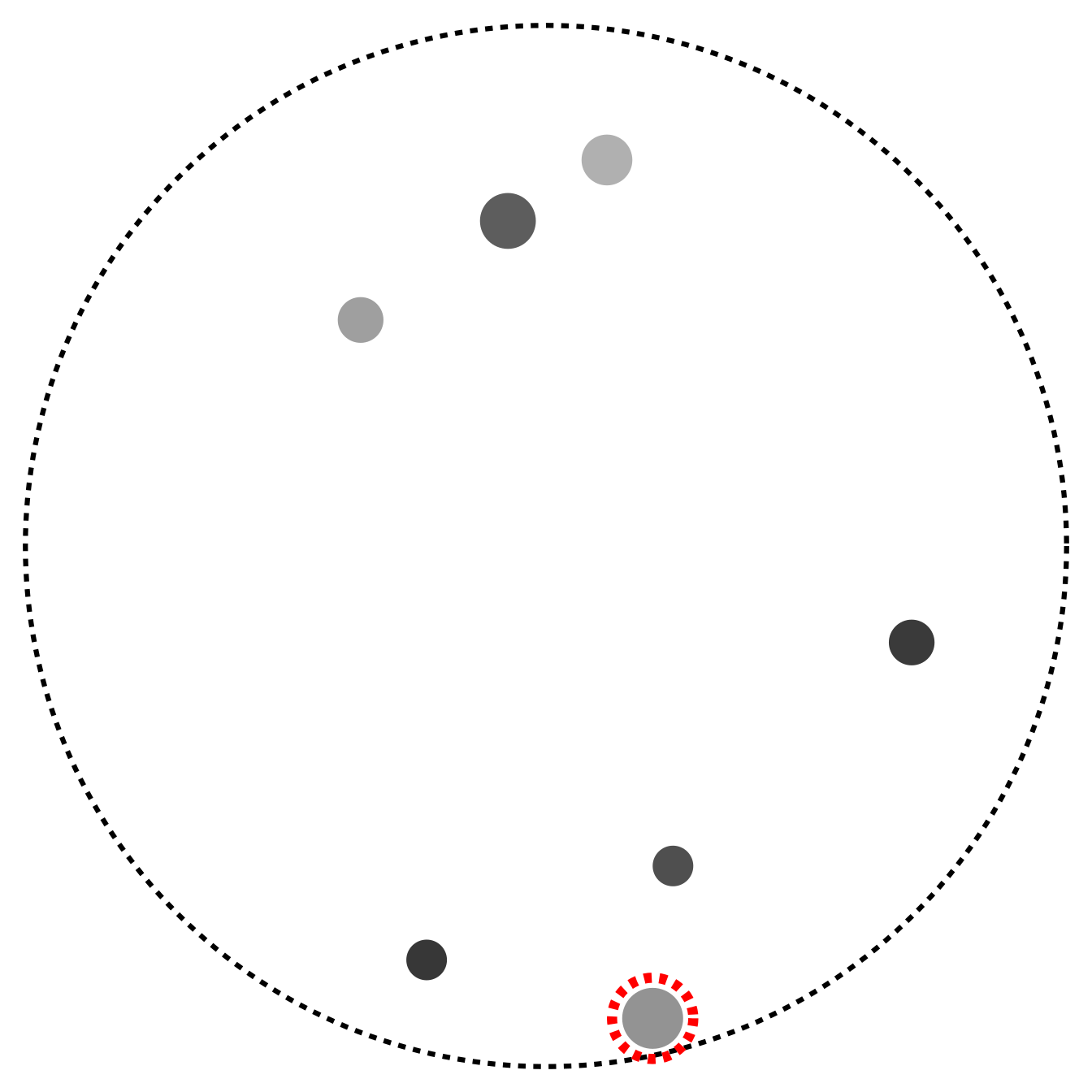}};
\node[inner sep=0pt] (agent_1) at (4.2,0)
  {\includegraphics[width=0.48\columnwidth]{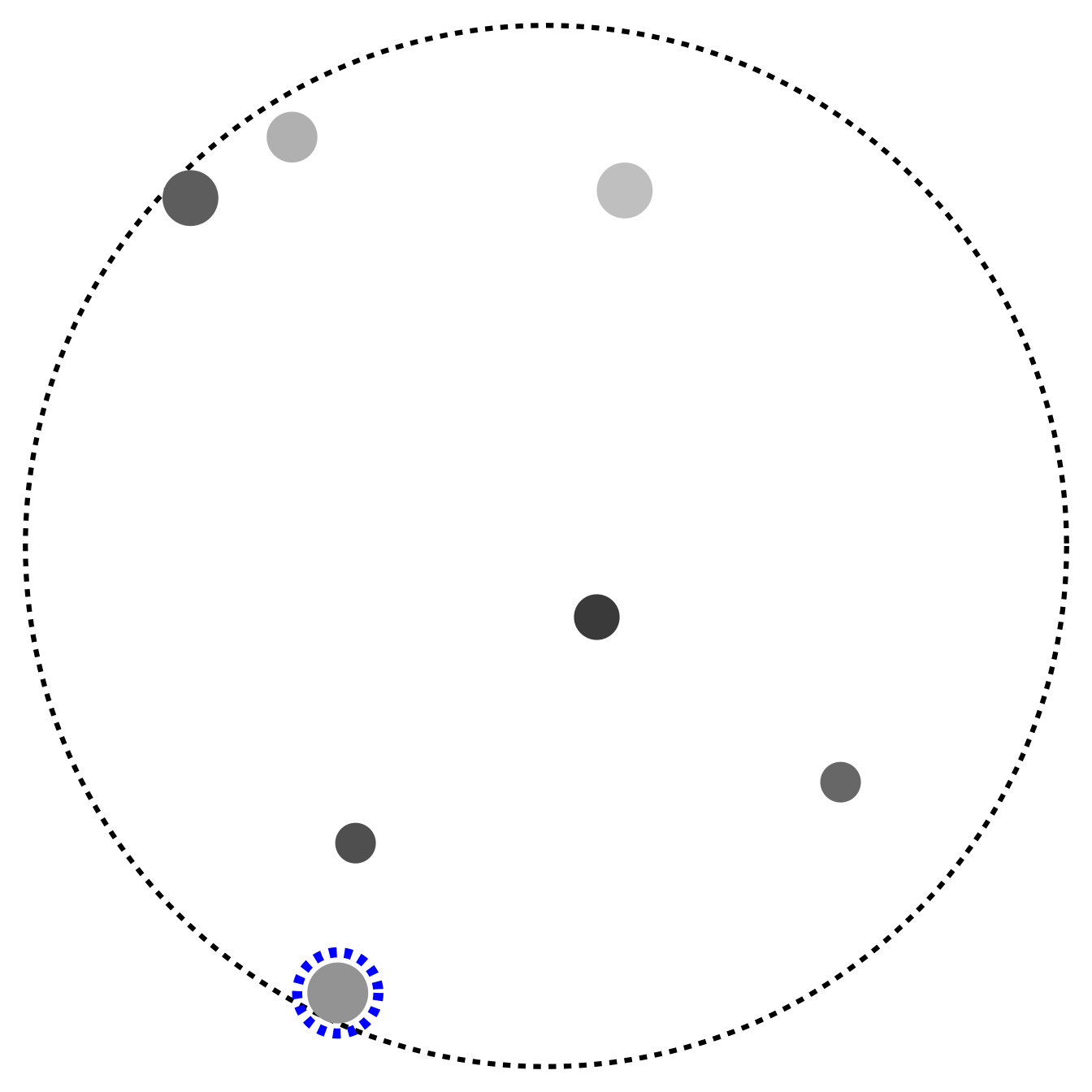}};
\node [below] at (0,-2) {A's view};
\node [below] at (4.2,-2) {B's view};
\end{tikzpicture}
\small
\begin{tabular}{@{}l@{}}
\toprule
A: I see three in a line going up and to the right. The middle one \\
is the largest and darkest \\
B: I don't see that. I have one large, medium gray dot that's under \\
a small, darker gray dot \\
A: Is the larger dot slightly to the left \\
B: yes, slightly, let's choose the larger one. \\
A: SELECT {\color{red} red} \\
B: SELECT {\color{blue} blue} \\
\bottomrule
\end{tabular}
\caption{Example dialogue and context of our task. Two agents have 7 entities in each view, but their views are centered slightly differently and only some entities are in common. Under this setting, agents must find a common entity through natural language communication.
}
\label{fig:example}
\end{figure}

One major goal of natural language processing is to develop agents with human-level competency in dialogue. In the field of human communication and development, the ability to construct and maintain common ground has been pointed out to be essential for natural language communication \cite{clark1996using} and acquisition \cite{tomasello2009constructing}. Furthermore, in the field of human-computer interaction, it is important that humans and computers have certain ways of creating \emph{mutual understandings} in order to collaboratively solve problems. Although natural language communication is not the only option, it is one of the most natural and effective solutions to this problem.

However, existing study of common grounding in dialogue system research is limited in three major ways.

First, existing dialogue tasks are limited in terms of common grounding due to the restricted types of information that need to be dealt with. Specifically, previous tasks are focused on either \emph{fully-observable} or \emph{categorical} context, which makes common grounding a relatively trivial task: 

\begin{itemize}
  \item In a \emph{fully-observable} context \cite{de2017guesswhat}, it is usually given that every information about the context is shared among the agents. This makes common grounding easier because information about the context is already in their common ground, and there could be little chance of misunderstandings. In contrast, under \emph{partially-observable} context agents typically need to create common ground from minimal shared information, and there could be a lot more misunderstandings between agents that need to be resolved.
  \item In a \emph{categorical} context \cite{bordes2016learning,he2017learning,lewis2017deal}, information can be expressed by symbolic natural language without ambiguity. For example, there could be little ambiguity in describing categorical properties, such as discrete color (\textit{red}, \textit{blue} and \textit{yellow}). However, in a \emph{continuous} context natural language usage can be ambiguous and more pragmatic (such as \textit{darker gray} and \textit{almost black}), and this introduces natural difficulty in terms of common grounding.
\end{itemize}

Another problem is the difficulty of evaluation and analysis. As the models acquire more flexibility, automatic evaluation becomes problematic \cite{liu2016,novikova2017} and interpretation of model behavior becomes more challenging. Since advanced common grounding requires high flexibility, it is expected that we will need reliable evaluation metrics and analysis methods in the process of comparing and improving different methods.

Finally, there are limitations of model capabilities. Although traditional dialogue systems rely on rule-based engineering and predetermined slot-filling \cite{traum1994computational,young2013pomdp,williams2016dialog}, these models lack flexibility in terms of representing dialogue states and generating natural utterances. Since common ground can be very complex which may include high ambiguity, uncertainty, partial understandings and misunderstandings, we need systems that can better capture such complexity and resolve them through flexible dialogues.

In this paper, we make a first step towards addressing these problems in the following way:

First, we formulate a novel dialogue task which requires advanced skills of common grounding under \emph{continuous} and \emph{partially-observable} context. Our task is based on a more general \emph{collaborative referring task}, where the goal of the agents is to coordinate attention on the same entity in a given context. This setting enables clear evaluation based on task success rate and various error analysis of complex models since the contexts are simple and completely controllable.

Second of all, to enable training of recent end-to-end dialogue systems with high flexibility \cite{bordes2016learning,lewis2017deal}, we collected a largescale dataset of 6,760 human dialogues with over 32K utterances through crowdsourcing on Amazon Mechanical Turk. During the dataset collection, we managed to fulfill three essential requirements of natural language corpora: interpretability, linguistic/strategic variety and reliability.

Next, we conduct a comparative analysis with the previous dataset to illustrate how continuous and partially-observable context introduces difficulty in terms of common grounding. In addition, further analyses of our dataset revealed various common grounding phenomena at different levels, including nonlinguistic bias towards \emph{joint saliency}.

Finally, we evaluate and analyze simple neural network models on our dataset based on an important subtask of collaborative referring. Due to the complexity of common grounding, there is still room for further improvement.

We show an example of the collected dialogue and context in Figure \ref{fig:example}. Although human players successfully coordinated the selection with relatively short turns, we can find difficult common grounding strategies such as pragmatic descriptions (\utterance{\textit{three in a line going up}}), clarification based on hypothesis testing (\utterance{\textit{Is the larger dot slightly to the left}}) and nuanced acknowledgement (\utterance{\textit{yes, slightly}}).

Overall, we expect our task to be a fundamental testbed for developing dialogue systems with advanced common grounding skills. Our dataset and scripts will be available at \url{https://github.com/Alab-NII/onecommon}.

\section{Related Work}
\label{section:related_work}

In dialogue system research, classical approach models common grounding based on finite states, where information in a dialogue transitions through fixed \emph{grounding acts} (such as \emph{Initiate, Continue, Repair, Request Repair, Acknowledge}) \cite{traum1994computational}, and the whole system relies on careful rule-based engineering and predetermined slot-filling \cite{young2013pomdp}. Although they perform reliably in restricted domains such as restaurant information retrieval \cite{williams2016dialog}, they lack flexibility to deal with realistic complexity of common ground.

Recently, a new line of data-driven dialogue systems, which we refer to as \emph{end-to-end dialogue systems}, has been gaining attention in both task-oriented \cite{bordes2016learning,lewis2017deal} and non-task-oriented domains \cite{vinyals2015neural,serban2016hierarchical}. In this approach, dialogue state and utterance generation are learned \emph{directly} from large raw corpora with little prior constraints, so they are more suitable for complex common grounding where flexibility is a requirement. However, few existing tasks focus on the difficulty of common grounding and most are based on either fully-observable or categorical context \cite{de2017guesswhat,bordes2016learning,lewis2017deal} where difficult common grounding is not required. A dataset closest to our setting is the MutualFriends dataset \cite{he2017learning}, which is based on the task of finding a mutual friend from private lists of friends. Although this can be considered as a collaborative referring task under \emph{partially-observable} context (due to the privacy of knowledge), they only include \emph{categorical} information and the difficulty of common grounding is limited. We give a precise comparative analysis in Section \ref{section:dataset_analysis}.

Referring expression generation and dialogues based on realistic visual context have also been studied extensively \cite{kazemzadeh2014referit,das2017visual,das2017visdial_rl}. Although these tasks are based on continuous (and sometimes partially-observable) context, realistic images could be too complex and costly to deal with modern dialogue systems. In contrast, we show that abstract and simple context is \emph{sufficient} for making common grounding difficult and \emph{preferable} for a testbed of common grounding where models can be easily developed, compared and analyzed. Nevertheless, we expect experiments on abstract and realistic contexts to have complementary strengths.

Finally, previous work addressed the difficulty of common grounding due to the \emph{perceptual difference} between humans and machines \cite{liu2012perceptual,fang2015embodied}. However, such problems are specific to human-machine dialogues, and instead we focus on a more general difficulty of common grounding due to complex ambiguity and uncertainty introduced by continuous and partially-observable context.

\section{Task Description}
\label{section:task_description}

\emph{Collaborative referring} is the task of creating a mutual understanding about the entity currently under discussion. We can interpret this as the initial step of common grounding where agents coordinate attention on specific entity (or entities) in the world, since it is only after collaborative referring succeeds that agents can exchange useful information about entities and develop common ground related to the world.

\subsection{Task Definition}

Formally, the task is formulated as a multi-agent cooperative game with two sets of entities $E = \{e_1, e_2, ... , e_m\}$ and agents $A = \{a_1, a_2, ... , a_n\}$. Each agent $a_i \in A$ has an observation of $E$, and at each timestep they can send natural language messages to coordinate their selections from $E$. The game is considered \emph{completed} when all the agents have made at least one selection action, and it is \emph{successful} if and only if all the agents selected the same entity.

We propose our task based on a minimal formulation of the collaborative referring task under \emph{continuous} and \emph{partially-observable} context. Specifically, we consider two agents located slightly differently in a 2-dimensional grid. Entities are also located in the same grid with two additional properties: \emph{size} and \emph{color}. However, agents can only observe entities within a fixed radius $r$ from their location, which makes this setting partially-observable. Furthermore, each observation is continuous, since all the properties of the entities (location, size and color) are continuous.

For the sake of simplicity, the number of entities observable by each agent is fixed at $7$. This ultimately reduces our task to a simple \textit{classification} problem, which can be evaluated based on simple metrics, such as accuracy.

\subsection{Dataset Collection}

\begin{figure}[ht]
\centering
\includegraphics[width=0.97\columnwidth]{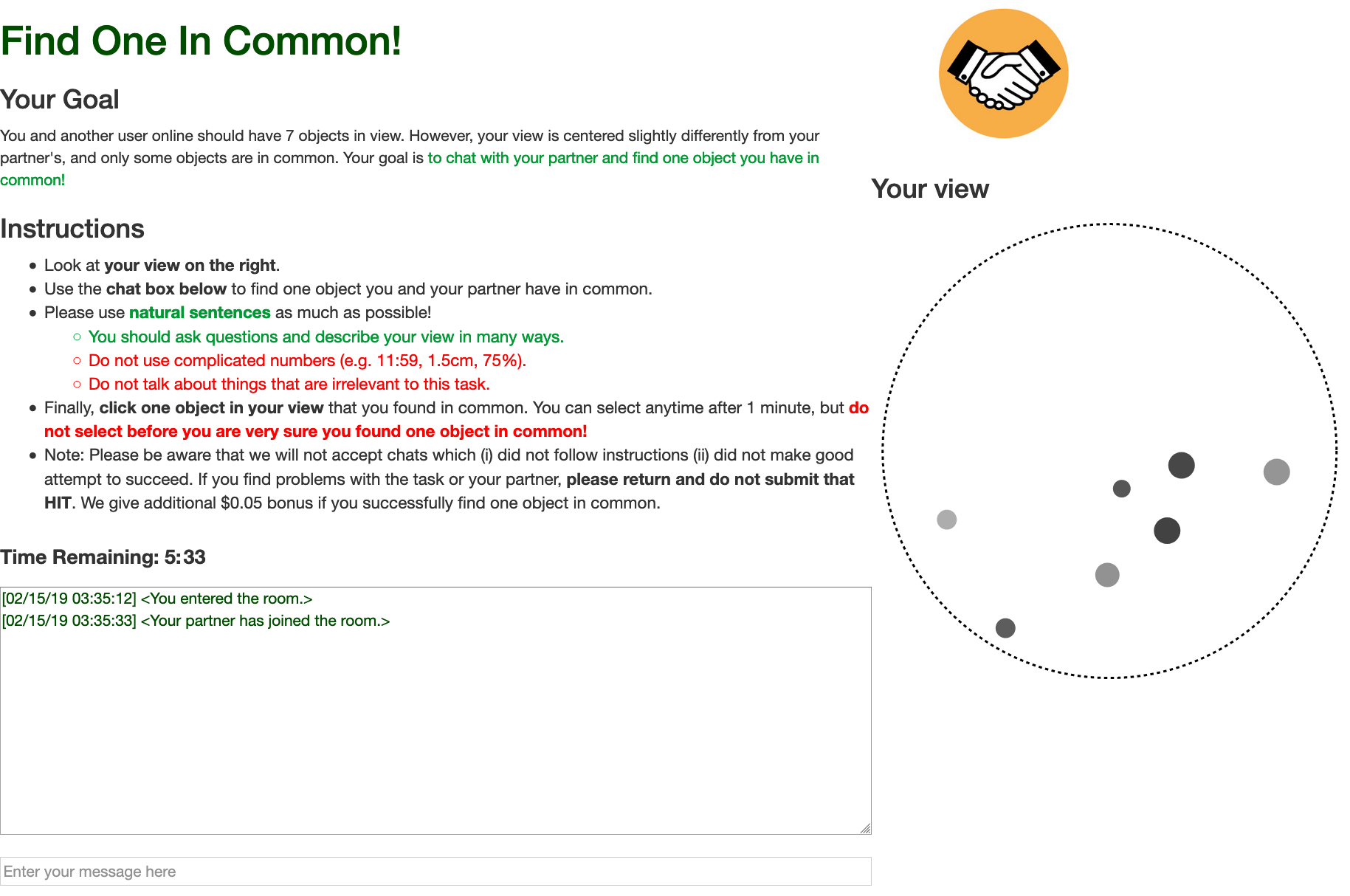}
\caption{
Screenshot of our chat interface. Workers are given a maximum of 6 minutes session to communicate through the chatbox and select the entity they found in common.
}
\label{fig:annotation}
\end{figure}

We basically followed the dataset collection procedure of the MutualFriends dataset \cite{he2017learning}. We used Amazon Mechanical Turk to pair up two human workers and gave each worker 20 seconds to read an example, followed by a maximum of 6 minutes session to complete the task. Our chat interface is shown in Figure \ref{fig:annotation}.

During the dataset collection, we were concerned on three essential requirements of natural language corpora: \emph{interpretability}, \emph{linguistic/strategic variety} and \emph{reliability}. In this section, we show why these properties need consideration and how we managed to fulfill these requirements.

\subsubsection{Interpretability}

We define the interpretability of the dataset to be the ease of interpreting its language and strategy, which is critical for additional annotations and error analysis. However, \emph{lack of discipline} and \emph{complexity of the vocabulary} can make a corpora difficult for interpretation.

In free-formed dialogues, lack of discipline can cause unnecessary difficulty in terms of interpretation. For instance, \emph{cross-talk} (conversation which does not progress sequentially) can occur frequently \cite{he2017learning} and complicate important structures of dialogues, such as discourse segments, adjacency pairs and \emph{contributions} in common grounding \cite{clark1996using}. Thus, we tried to minimize them by forcing workers to take turns. Also, \emph{chit-chat} could occur occasionally, which adds undesirable noise for analyzing strategies. Therefore, we explicitly instructed the workers to avoid talking about things that are irrelevant to this task.

Keeping the vocabulary simple is also important for interpretability, especially for people unfamiliar with the domain. For instance, the MutualFriends dataset \cite{he2017learning} includes up to 7 attributes with approximately 3K complex named entities and technical terms. In contrast, we kept the attributes minimal with only 4 intuitive scalar attributes (x-value, y-value, size and color). As a result, this greatly reduced the complexity of the vocabulary as we describe in detail in Section \ref{section:dataset_analysis}.

\subsubsection{Linguistic and Strategic Variety}

\emph{Linguistic and strategic variety} of the dataset is fundamental for developing dialogue systems with broad coverage. To elicit this property, we sampled all attributes of the entities uniformly at random, with the only restriction that the entities cannot be too close to each other. As the previous work with similar idea confirmed \cite{suhr2017corpus}, we found rich varieties of linguistic phenomena, including cardinalities (\utterance{\textit{three} gray dots}), existentials (\utterance{\textit{There is} another small dark ..}), universals (\utterance{\textit{all} of the other dots are larger}), coordinations and negations (\utterance{further to the right \textit{and} \textit{not} as far down}).

During the dataset collection, we assigned 6,759 unique contexts to 6,760 dialogues we collected. We collected two dialogues based on the exact same context and confirmed that they solved the task in different ways: thus there could be various effective solutions and agents must adapt flexibly to their partners' strategies.

In addition, we gave variation according to the \textit{degree} of partial observability. Specifically, only 4, 5 or 6 out of the 7 entities in view are shared among the agents. As a result, we found further variation of common grounding strategies, as we discuss in detail in Section \ref{section:dataset_analysis}.

\subsubsection{Reliability}

Finally, we regard the \emph{reliability} of the dataset to be crucial, especially when crowdsourcing data through untrained, possibly low-motivated workers. Specifically, in our preliminary experiment we found many cases where workers did not follow the instruction carefully or solve the task effectively, especially on difficult cases.

As a solution, we manually reviewed all work and rejected ones which clearly did not follow the instruction. Our instruction is kept brief and explicit so that it is easier to follow, and we also gave manual feedback about general solutions to improve their work. To discourage premature guessing, we prohibited workers from selecting within the first minute and instructed them to make it \emph{very sure} they found one entity in common. We also incentivized task success with \$0.05 bonus for all successful dialogues, in addition to the base reward of \$0.30.

As a result, we found significant improvement in terms of task success rate, which is an important evidence of the reliability of our dataset. \\

Based on the above procedure, we collected 6,839 completed dialogues and accepted 6,760 dialogues in total. Overall, we received positive feedback about our task and the workers seemed to enjoy it.

\section{Dataset Analysis}
\label{section:dataset_analysis}

In this section, we verify the difficulty of common grounding introduced in our task and conduct further analyses of our dataset.

\subsection{Difficulty of Common Grounding}

Our hypothesis is that \emph{continuous} and \emph{partially-observable} context makes common grounding difficult compared to \emph{categorical} or \emph{fully-observable} context. Since our task can be solved trivially with full-observability (e.g., by always uttering \utterance{\textit{select the darkest dot}}), we focus on testing how \textit{continuous} context adds difficulty in terms of common grounding.

As a comparison, we use the MutualFriends dataset which is based on a similar collaborative referring task under partially-observable but \emph{categorical} context \cite{he2017learning}. However, several differences make a direct comparison difficult: for instance, we only gave one chance for final selection, while the MutualFriends dataset allowed multiple chances in the given time. Therefore, we focused on the following factors which are less affected by the differences.

\begin{table*}[ht]
\centering
\begin{tabular}{c|cccc}
\toprule

 & \multirow{2}{*}{MutualFriends} & \multicolumn{3}{c}{Ours} \\
 & & (\# Shared = $4$) & (\# Shared = $5$) & (\# Shared = $6$) \\
\midrule
\# dialogues &
10,661 & 2,189 & 2,279 & 2,292 \\
Average tokens per utterance&
5.38 & 12.87 & 12.37 & 11.86\\
Average turns per dialogue&
8.97$^*$ & 4.97 & 4.77 & 4.56 \\
Success rate &
0.85$^*$ & 0.66 & 0.77 & 0.87 \\
\midrule
Unique tokens & 13,478 & \multicolumn{3}{c}{3,761} \\
Occupancy of top 10\% frequent tokens & 91.6\% & \multicolumn{3}{c}{97.0\%} \\
\bottomrule
\end{tabular}
\caption{\label{statistics}
Basic statistics of our dataset and the MutualFriends dataset \cite{he2017learning}. To count tokens and vocabulary size, we preprocessed the text with the same NLTK word tokenizer \cite{nltkbook} and converted each token to its lowercased form. Statistics with asterisk are not suitable for direct comparison due to the task difference (e.g. the number of chances).}
\end{table*}

\begin{table*}[ht]
\centering
\begin{tabular}{l|cc|c|l}
\toprule
Nuance Type & MutualFriends & Ours & Example Keywords & Example Usage \\
\midrule
Approximation (10) & 0.10 & 3.98 & almost, nearly, approximately & \textbf{almost} in the middle \\
Exactness/Confidence (33) & 0.14 & 2.71 & exactly, completely, definitely & \textbf{exactly} horizontal \\
Subtlety (12) & 0.01 & 9.37 & slightly, bit, somewhat & \textbf{slightly} to the right \\
Extremity (27) & 0.21 & 9.35 & very, really, extraordinary & \textbf{very} light dot \\
Uncertainty (20) & 0.57 & 5.79 & maybe, might, guess & \textbf{Maybe} it's different \\
\bottomrule
\end{tabular}
\caption{\label{nuances}
Average occurence of nuanced expressions per 100 utterances (dictionary size shown in parenthesis).
}
\end{table*}

\subsubsection{Utterance Length}

We compare the \emph{average utterance length} because this indicates the syntactical/semantic complexity of utterances required for common grounding. As shown in Table \ref{statistics}, utterances in our dataset are at least twice as long as those in the MutualFriends dataset. Therefore, more complex utterances are required under continuous context compared to categorical context. Interestingly, utterance lengths also slightly increase when the number of shared entities are smaller: thus, greater \emph{degrees} of partial-observability also add complexity at the utterance level.

\subsubsection{Pragmatic Expressions}

\begin{figure}[ht]
\centering
\begin{tikzpicture}
\node[inner sep=0pt] (agent_0) at (0,0)
  {\includegraphics[width=0.48\columnwidth]{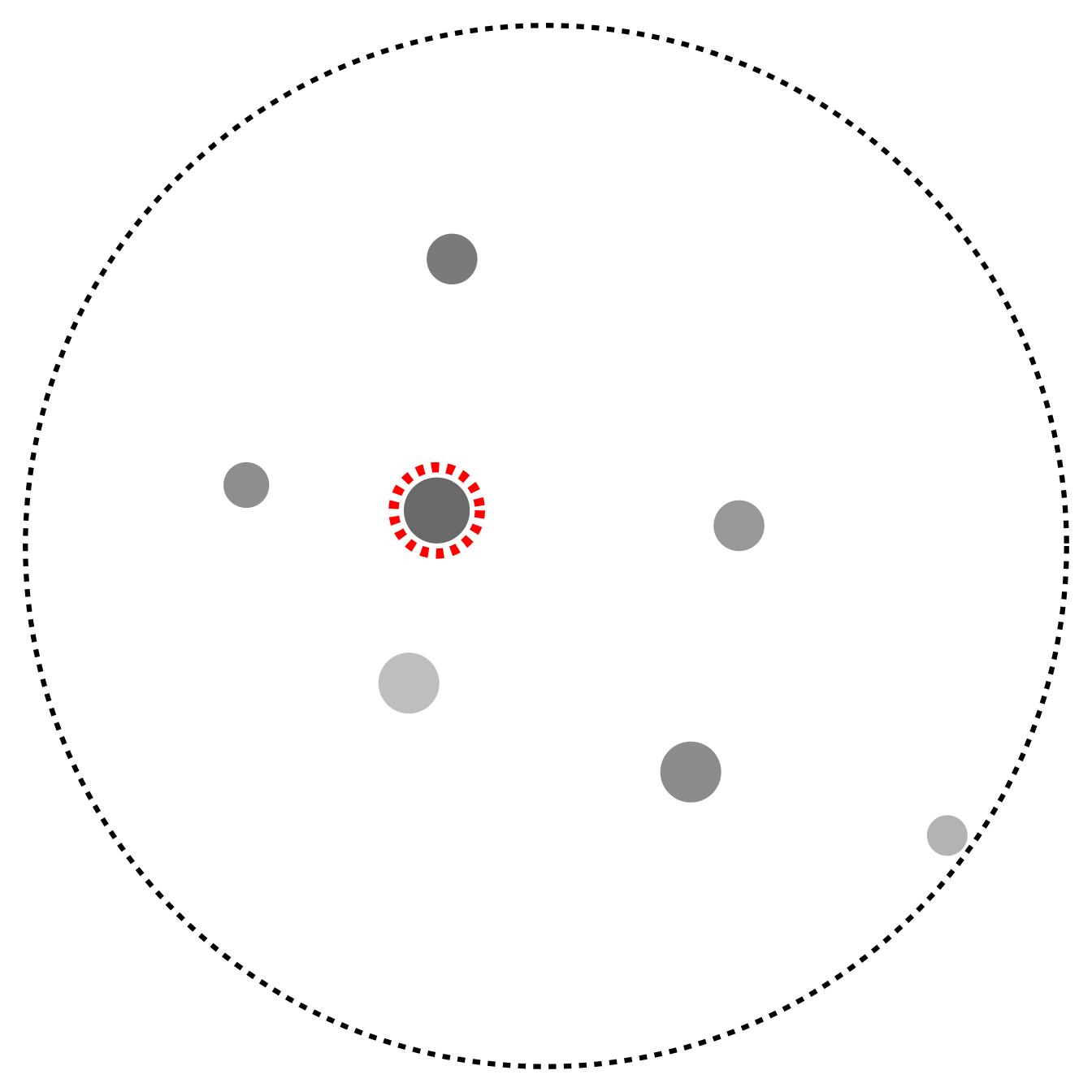}};
\node[inner sep=0pt] (agent_1) at (4.2,0)
  {\includegraphics[width=0.48\columnwidth]{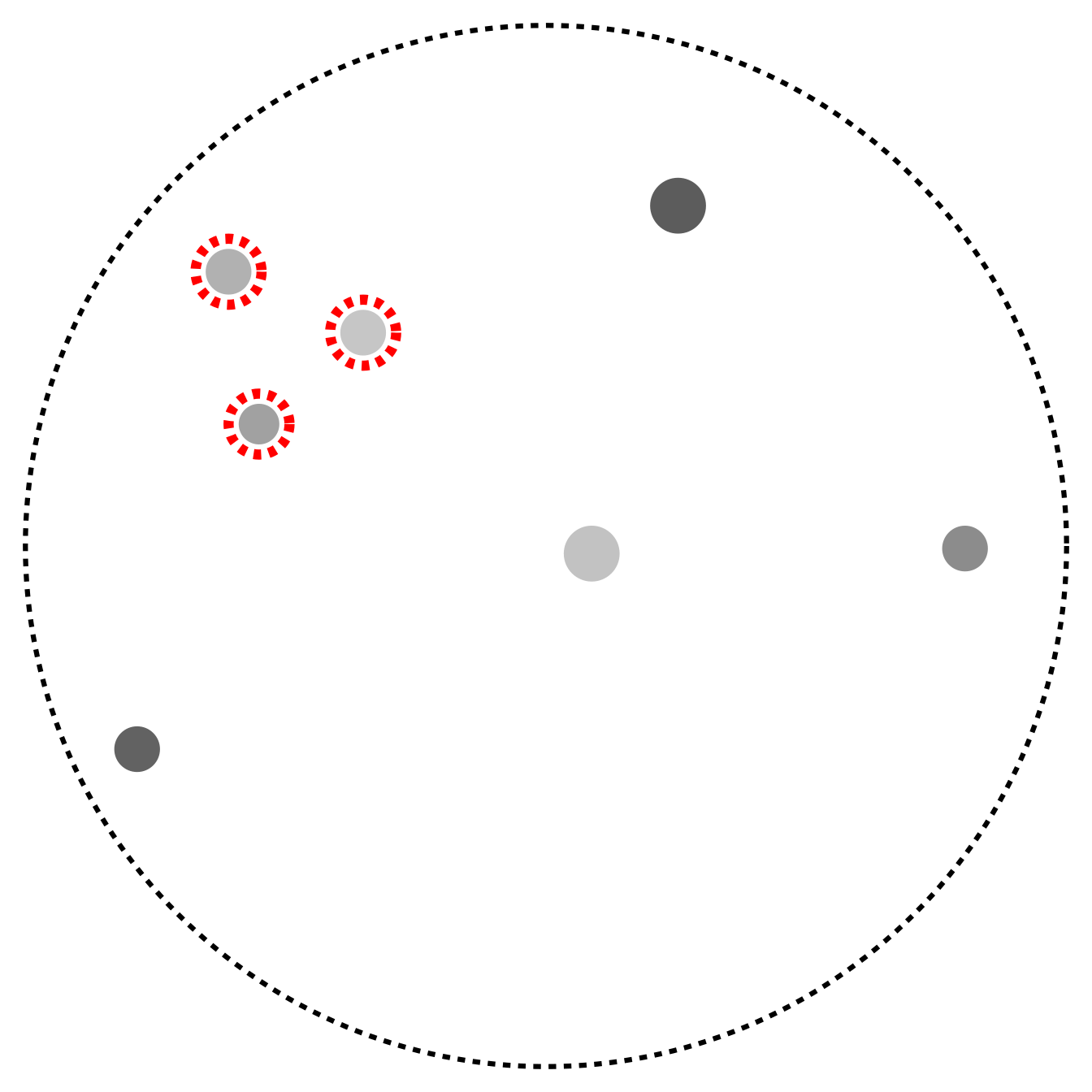}};
\node [below] at (0,-2) {large \textbf{black} dot};
\node [below] at (4.2,-2) {a close \textbf{triangle}};
\end{tikzpicture}
\caption{
Examples of typical pragmatic expressions in our dataset, marked by bold.
}
\label{fig:pragmatics}
\end{figure}

In our dataset, we found many \emph{pragmatic expressions} whose meaning depend on the context and should not be taken literally. A typical example is the usage of word \textit{black} to indicate the darkest dot in the context, even if its color is not completely black. Another common expression \textit{triangle} is also pragmatic, since in literal sense there could be numerous triangles in one's view, and the speaker actually indicates a group of three dots which is close to prototypical types of triangles, such as an equilateral triangle. Examples are shown in Figure \ref{fig:pragmatics}.

As the previous work pointed out, such pragmatic expressions are characteristic in continuous context \cite{monroe2017colors} and add complex ambiguity that need to be resolved through common grounding.

\subsubsection{Nuanced Expressions}

Finally, frequent usage of \emph{nuanced} expressions is an important characteristic of our dataset. Since the context is continuous and partially-observable, we hypothesize that speakers need to rely on such expressions to express subtle differences in terms of degree, ambiguity and uncertainty. 

To estimate the frequency of nuanced expressions, we manually reviewed 100 dialogues from each dataset to create keyword-based dictionaries of nuanced expressions, which are further expanded with synonyms/morphologies. We excluded words which are likely to be used with different meanings (such as \textit{like}, \textit{about} and \textit{around}). For simplicity, we do not consider nuances expressed morphologically (such as \textit{ish} as in \textit{smallish}) although they are also common in our dataset. Results in Table \ref{nuances} show that our dataset includes significantly more nuanced expressions of various types. The dictionaries will be publicly available for reproducibility. \\

To summarize the points, utterances are much longer in our dataset, which indicates the complexity of common grounding at the utterance level. Secondly, our dataset includes more ambiguity and uncertainty represented by frequent occurrence of pragmatic expressions and nuanced expressions. Thus the process of common grounding is complicated, but regardless of such difficulties, human workers could solve the task reasonably well with little evidence of confusion. Therefore, we conclude that introducing continuity and partially-observability to the context is a critical solution to add natural difficulty in terms of common grounding.

\subsection{Further Analysis}

Next, we conduct further analyses of the collected dataset, which revealed important phenomena at different levels that need to be considered.

\subsubsection{Basic Statistics}

From Table \ref{statistics}, we also found that dialogues get longer in terms of \emph{average turn length} with fewer shared entities. This shows that under greater degrees of partial-observability, it is more likely that the presented information is not \emph{groundable} and players need more try-and-error to create common ground. Success rate also drops naturally, so in general common grounding becomes more challenging when less information is shared.

In terms of lexical variety, we found 3,761 unique tokens in total, in contrast to 13,478 in the MutualFriends dataset. In addition, large portion of the corpora constitutes of common words, and top 10\% of the most frequent tokens occupy 97.0\% of the whole tokenized corpora, in contrast to 91.6\% in the MutualFriends dataset. Therefore the vocabulary of our dataset is extremely simple, which is an important evidence of interpretability discussed in Section \ref{section:task_description}. This may also be helpful for training dialogue systems because rare words are less problematic.

\subsubsection{Nonlinguistic Phenomena}

Language is a \emph{coordination device} we use to coordinate our joint actions \cite{lewis1969convention}, but we also use \emph{joint saliency} to coordinate actions at the nonlinguistic level \cite{schelling1980strategy}. In our dataset, we found that human players have a tendency to focus attention on \emph{perceptually} salient entities more often.

\begin{figure}[ht]
\centering
\begin{tikzpicture}
\node[inner sep=0pt] (popup) at (0,0)
  {\includegraphics[width=0.94\columnwidth]{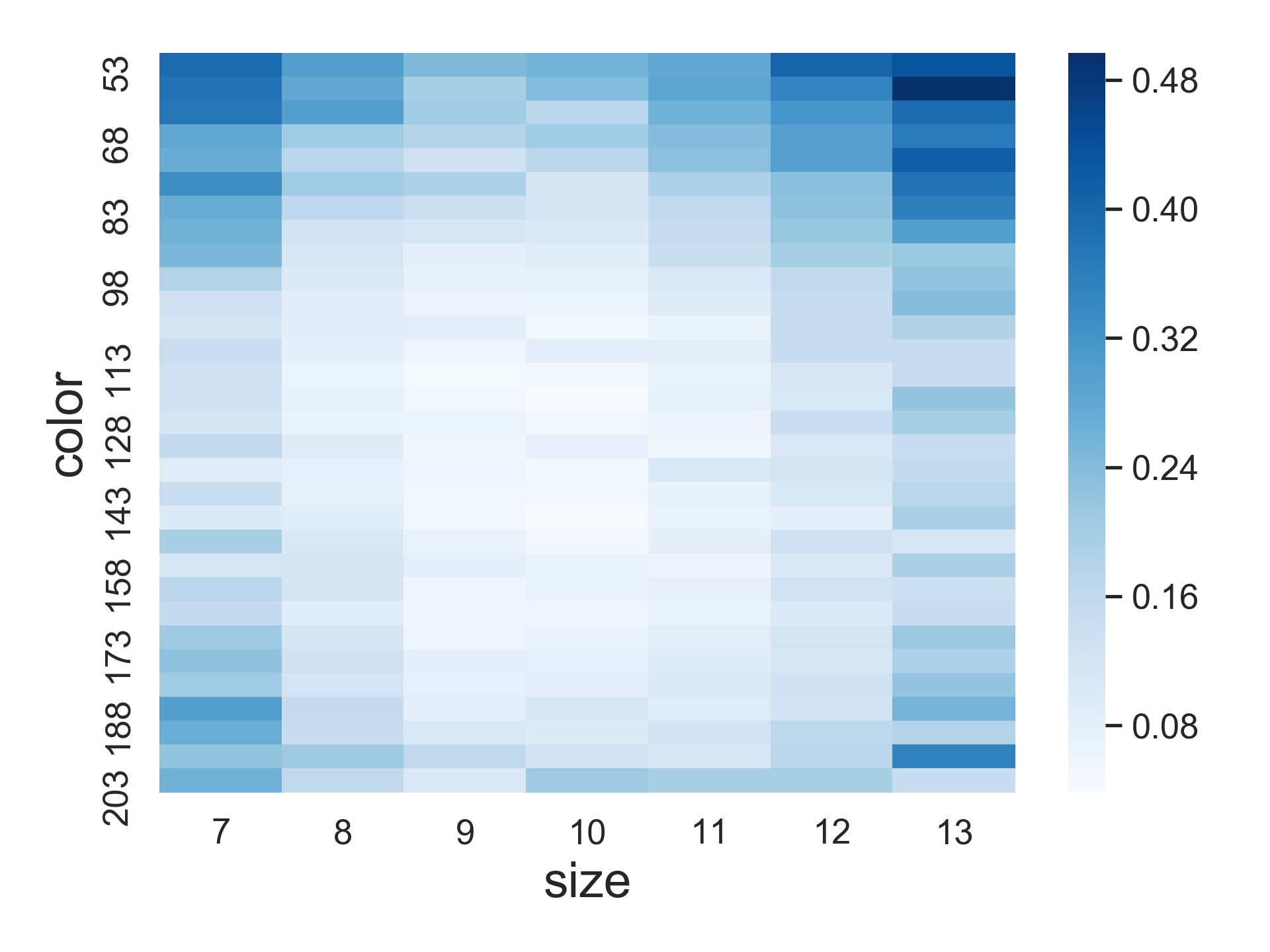}};
\end{tikzpicture}
\caption{
Comparison of the final selection probabilities based on color and size. The total range of size is 7 and color is split into 30 equal-sized bins based on RGB scale (smaller is darker).
}
\label{fig:plot-select}
\end{figure}

We plot the final selection probabilities based on entity's color and size in Figure \ref{fig:plot-select}. We can clearly see that the selection is biased, and entities with extreme properties (around the edge) are more likely to be selected. We also found that darker entities are more likely to be selected (62.7\%) compared to lighter entities (37.3\%), and larger entities (54.3\%) slightly more likely than smaller entities (45.7\%).

There could be other types of joint saliency such as geometric relations between entities, but the point is that such bias exists and needs consideration: for example, just by taking advantage of such bias, we can predict human selections significantly better than random (Section \ref{section:experiments}). However, due to partial-observability joint saliency is not sufficient to solve our task and communication is critical.

\begin{table*}[htb]
\centering
\begin{tabular}{lcl}
\toprule
Function & Type & Example Utterances\\
\midrule
\multirow{8}{*}{Information Providing} & \multirow{1}{*}{Inform (Init.)} & I have very dark small dot in the center \\
& \multirow{1}{*}{Inform (Cont.)} & It also has a small light grey one further down from the group \\
& \multirow{1}{*}{Agreement} & Yes I have one like that. / same here. \\
& \multirow{1}{*}{Agreement (Strong)} & Exactly! / perfect. mine too. \\
& \multirow{2}{*}{Agreement (Partial)} & not sure its the one / more of a line. \\
& & Yes, but the small is medium dark, not completely black \\
& \multirow{1}{*}{Disagreement} & I don't have that one. / mine are not in those locations. \\
\midrule
\multirow{4}{*}{Information Seeking} & \multirow{1}{*}{Question (Prop.)} & the middle one is the darkest of the 3? \\
& \multirow{1}{*}{Question (Set)} & where is it in relation to the large med grey? \\
& \multirow{1}{*}{Question (Choice)} & Which should we choose? / the black or the grey? \\
& \multirow{1}{*}{Question (Check)} & It's the darkest dot in the circle, right? \\
\midrule
\multirow{1}{*}{Commissives} & \multirow{1}{*}{Offer} & lets click the upper left one that's bigger and darker gray \\
\midrule
\multirow{2}{*}{Directives} & \multirow{1}{*}{Request} & tell me about your tiniest dot? / pick one at the bottom \\
& \multirow{1}{*}{Suggestion} & Please describe it in relation to other dots in the circle \\
\bottomrule
\end{tabular}
\caption{\label{speechact}
Illustrative utterances in the dataset, grouped by the \emph{task dimension} of communicative functions \cite{bunt2017dialogue}}
\end{table*}

\subsubsection{Utterance Level Phenomena}

Understanding speaker's meaning at the \emph{utterance level} is critical in dialogue \cite{grice1957meaning}: especially the idea of speech acts \cite{austin1975things,searle1969speech} has been applied widely in dialogue system research for improving utterance understanding and generation.

In our setting, we allowed free-formed chat with few restrictions, as long as they are relevant for the accomplishment of the task: as a result, we found a wide variety of speech acts in various forms related to common grounding. We show illustrative examples of the collected utterances in Table \ref{speechact}. Utterances are grouped by the \emph{task dimension} of communicative functions \cite{bunt2017dialogue}, including information transfer functions (Information Providing/Seeking) and action discussion functions (Commissives/Directives). With additional annotations, our dataset can be extended for other dialogue tasks, such as dialogue act recognition.

\subsubsection{Discourse Level Phenomena}

Naturally, we found many coreference and anaphoric expressions in our dataset. \emph{Coreference resolution} is the task of mapping mentions of entities to their referents. In our dataset, we found two characteristics that complicate this task. First, due to continuous and partially-observable context, mentions are usually ambiguous and referents may be missing. Thus players must keep track of various possibilities and investigate them through interaction. Secondly, players often use \emph{groupings} (such as \textit{three in a line}, \textit{a cluster of 4 dots}) where mentions refer to \emph{sets} of entities. This strategy could be effective but adds complexity to coreference resolution.

On the other hand, \emph{anaphoric relation} is the relation between a mention and following mentions which refer to the previous mention. This can occur both within utterances (\utterance{a medium size black one, with a very light slightly smaller one to \textit{it's} left}) and across utterances (\utterance{Does \textit{the lighter dot} appear to be slightly larger?}). Similar to coreference resolution, this is a challenging subtask of common grounding at the discourse level which can be studied on our dataset.

\section{Experiments}
\label{section:experiments}

\subsection{Experiment Overview}

In this experiment, we formulate a natural language understanding task based on \textit{target selection}: specifically, we try to predict which target a player selected, given the player's observation and the corresponding dialogue. This is an essential subtask of collaborative referring, where players choose their final selection based on the created common ground. Since the number of entities in view is fixed at 7, we can formulate this as a simple classification problem. Our baseline models are kept as simple as possible, with minimal preprocessing and hyperparameter tuning.

\subsection{Methods}

\begin{table*}[htb!]
\centering \small
\begin{tabular}{c|ccc}
\toprule
 & Full & Uncorrelated & Success Only \\
\midrule
Random & 14.28 & 14.28 & 14.28 \\
\midrule
Context Only (MLP) & 27.90 $\pm$ 0.6 & 28.74 & 29.59 \\
Context Only (RN) & 31.94 $\pm$ 0.9 & 30.22 & 32.40 \\
Context + Dialogue (MLP) & 40.27 $\pm$ 1.3 & 40.89 & 43.82 \\
Context + Dialogue (RN) & 43.09 $\pm$ 0.8 & 44.00 & 49.44 \\
\midrule
Humans & - & 82.50 & 90.79 \\
\bottomrule
\end{tabular}
\caption{\label{selection_experiment}
Results of the target selection experiment. Models are trained 10 times initialized with different seeds for the Full testset, and the models with best validation loss are used for the additional testset results (Uncorrelated and Success Only).
}
\end{table*}

Two main components of the models are as follows:

\subsubsection{Context Embedding}
The structured form of the context is represented as a 28 dimensional real-valued vector, where each of the 7 observable entities is represented as a 4 dimensional vector (x-value, y-value, size, color). Each dimension is further normalized in the range of (-1,1).

The simplest way to embed context is to directly apply a multi-layered perceptron (MLP) over the context vector. However, without feature engineering this simple approach may have difficultly in capturing relevant information, such as relations between entities. Therefore, in the second approach we use the Relation Network module \cite{santoro2017simple} to create additional features about relations between entities. Specifically, we embed each combination of the entities (total of 21 pairs) with a shared MLP and append the sum of these vectors as additional input.

\subsubsection{Dialogue Embedding}

Utterances are all tokenized and lowercased, and tokens which occur less than 10 times are treated as a unique \emph{unknown} token. We insert tokens which represent \emph{speaker id} to each utterance at the beginning, and another token to indicate the end of the dialogue. Then, we embed these tokens with a shared MLP and run a bidirectional GRU \cite{cho2014properties} over the embedded tokens. Finally, we take the last output of the bi-GRU as the final representation of the dialogue. \\

For prediction, we simply concatenate the context and dialogue embeddings and run another MLP. However, as we've seen in Section \ref{section:dataset_analysis}, there are nonlinguistic selection bias in our dataset, so it is possible to make predictions without dialogue embeddings. Therefore, we also train models to predict only from the context embeddings using MLP.

Following common practice, we split the dataset into training, validation and test set with a proportion of 8:1:1, and all models are tuned on the validation set. The loss function is calculated using cross entropy. All components of the neural networks consist of single layer with 128 hidden units, and dropout rate of 0.5 is applied at each layer to avoid overfitting. All parameters are initialized uniformly within the range of (-0.01, 0.01). Models are trained with the Adam optimizer \cite{kingma2014adam} with initial learning rate of 0.001, and we clip gradients whose $L^2$ norm is greater than 0.1. The experiment is run 10 times initialized with different seeds, and we report the mean and standard deviation of the selection accuracies on the full testset.

For further analyses, models with the best validation loss in the previous experiment are also tested on two variants of the testset. First, we create an uncorrelated testset by randomly removing one from each correlated pair in the current testset (same dialogue but different context). Secondly, we further removed dialogues where players failed to coordinate on the same entity from the uncorrelated testset, since this may affect target selection performance. The statistical significance of the results for each pair of methods are tested on the uncorrelated testset using paired student's t-test. Finally, we take 100 random samples from the uncorrelated testset (including 76 successful) to report human performance based on average accuracy of two annotators.

\subsection{Results}

We show the results of our experiment in Table \ref{selection_experiment}. As we can see, models trained only with the context embeddings perform significantly better than random (\textit{p}-value $<10^{-7}$). This verifies that we can indeed take advantage of selection bias to make better predictions.

In addition, we found that embedding context with Relation Network consistently outperforms MLP, but not at a statistically significant level (\textit{p}-value $>0.1$). Therefore, the simplest strategy of using MLP works decently, but a better architecture may improve the overall performance.

Finally, models trained with both context and dialogue embeddings significantly outperform models trained only with the context embeddings (\textit{p}-value $<10^{-9}$). This indicates that even our simplest models can learn to ground linguistic meanings based on the context to make better predictions. When the testset only includes successful cases, models perform better but human performance improves even more achieving over 90\% accuracy. Overall, our target selection task is challenging due to the complexity of common grounding, and we still have a huge room for improvement.

\section{Conclusion and Future Work}
\label{section:conclusion}

The main contributions can be summarized as follows:

\begin{itemize}
  \item We proposed a simple and general idea of incorporating continuous and partially-observable context to the dialogue tasks, which makes common grounding difficult in a natural way.
  \item Following this idea, we formulated a novel dialogue task based on collaborative referring which enables clear evaluation and analysis of complex models.
  \item We collected a largescale dataset of 6,760 dialogues, which fulfills essential requirements of natural language corpora and will be publicly available online.
  \item Our analysis of the dataset verified the difficulty of common grounding and revealed various phenomena that need to be considered.
  \item We evaluated and analyzed simple baseline models on an important subtask of collaborative referring and showed that there is still room for further improvement.
\end{itemize}

In future work, we will evaluate and analyze dialogue models based on our task, especially to identify the current limitations of end-to-end approaches in terms of common grounding. Models can be trained in a variety of ways, including supervised learning, reinforcement learning with humans, and reinforcement learning based on \emph{self-play} \cite{lewis2017deal}. Overall, we expect our task to be a fundamental testbed for developing dialogue systems with sophisticated common grounding abilities.

\section*{Acknowledgements}
This work was supported by JSPS KAKENHI Grant Numbers 16K12546,18H03297.

\fontsize{9.5pt}{10.5pt} \selectfont

\bibliography{myref}
\bibliographystyle{aaai}

\end{document}